\documentclass[10pt,twocolumn,letterpaper]{article}

\usepackage{cvpr}              
\usepackage{bbm}
\usepackage{multirow}
\usepackage{colortbl}
\usepackage{pifont}
\usepackage{marvosym}
\usepackage[accsupp]{axessibility} 
\definecolor{HeaderGray}{RGB}{235, 235, 240}    
\definecolor{LightBlueRow}{RGB}{248, 252, 255}  
\definecolor{OursHighlight}{RGB}{245, 240, 255} 
\definecolor{DeepBlueText}{RGB}{0, 70, 140}     
\definecolor{cvprblue}{rgb}{0.21,0.49,0.74}
\usepackage[pagebackref,breaklinks,colorlinks,allcolors=cvprblue]{hyperref}

\def\paperID{*****} 
\def\confName{CVPR}
\def\confYear{2026}
\def\ourmodel{PF-RPN}
\graphicspath{{./figure/}}
\DeclareGraphicsExtensions{.jpg,.pdf,.png}

\title{Prompt-Free Universal Region Proposal Network}

\author{
Qihong Tang\textsuperscript{\rm 1}\thanks{Equal Contribution, \\ \indent$\textsuperscript{\Letter}$Correspondence to: Qi Fan \texttt{<fanqi@nju.edu.cn>}.}\;,
Changhan Liu\textsuperscript{\rm 1}$^*$,
Shaofeng Zhang\textsuperscript{\rm2},
Wenbin Li\textsuperscript{\rm 1},
Qi Fan\textsuperscript{\rm 1 \Letter},
Yang Gao\textsuperscript{\rm 1} \\[2mm]
$^1$Nanjing University, $^2$University of Science and Technology of China \\[2mm]
}

\begin{document}
\maketitle
\makeatletter
\makeatother
\begin{abstract}
Identifying potential objects is critical for object recognition and analysis across various computer vision applications.
Existing methods typically localize potential objects by relying on exemplar images, predefined categories, or textual descriptions.
However, their reliance on image and text prompts often limits flexibility, restricting adaptability in real-world scenarios.
In this paper, we introduce a novel Prompt-Free Universal Region Proposal Network (\ourmodel), which identifies potential objects without relying on external prompts.
First, the Sparse Image-Aware Adapter (SIA) module performs initial localization of potential objects using a learnable query embedding dynamically updated with visual features.
Next, the Cascade Self-Prompt (CSP) module identifies the remaining potential objects by leveraging the self-prompted learnable embedding, autonomously aggregating informative visual features in a cascading manner.
Finally, the Centerness-Guided Query Selection (CG-QS) module facilitates the selection of high-quality query embeddings using a centerness scoring network.
Our method can be optimized with limited data (e.g., 5\% of MS COCO data) and applied directly to various object detection application domains for identifying potential objects without fine-tuning, such as underwater object detection, industrial defect detection, and remote sensing image object detection.
Experimental results across 19 datasets validate the effectiveness of our method. Code is available at \href{https://github.com/tangqh03/PF-RPN}{https://github.com/tangqh03/PF-RPN}.

\end{abstract}    
\section{Introduction}
\label{sec:intro}
Recent object detection methods with Region Proposal Networks (RPN)~\cite{xu2023devit,fu2024cross} have achieved significant progress in various computer vision applications.
The Region Proposal Network (RPN) generates a sparse set of proposal boxes for potential objects, which is a key component of object detection.
However, existing RPN methods~\cite{ren2016faster,vu2019cascade,zou2021sc} often fail to identify potential target objects from unseen domains.
This limitation significantly hinders object detection application in open-world scenarios.

\begin{figure}[!t]
    \centering
    \includegraphics[width=\linewidth]{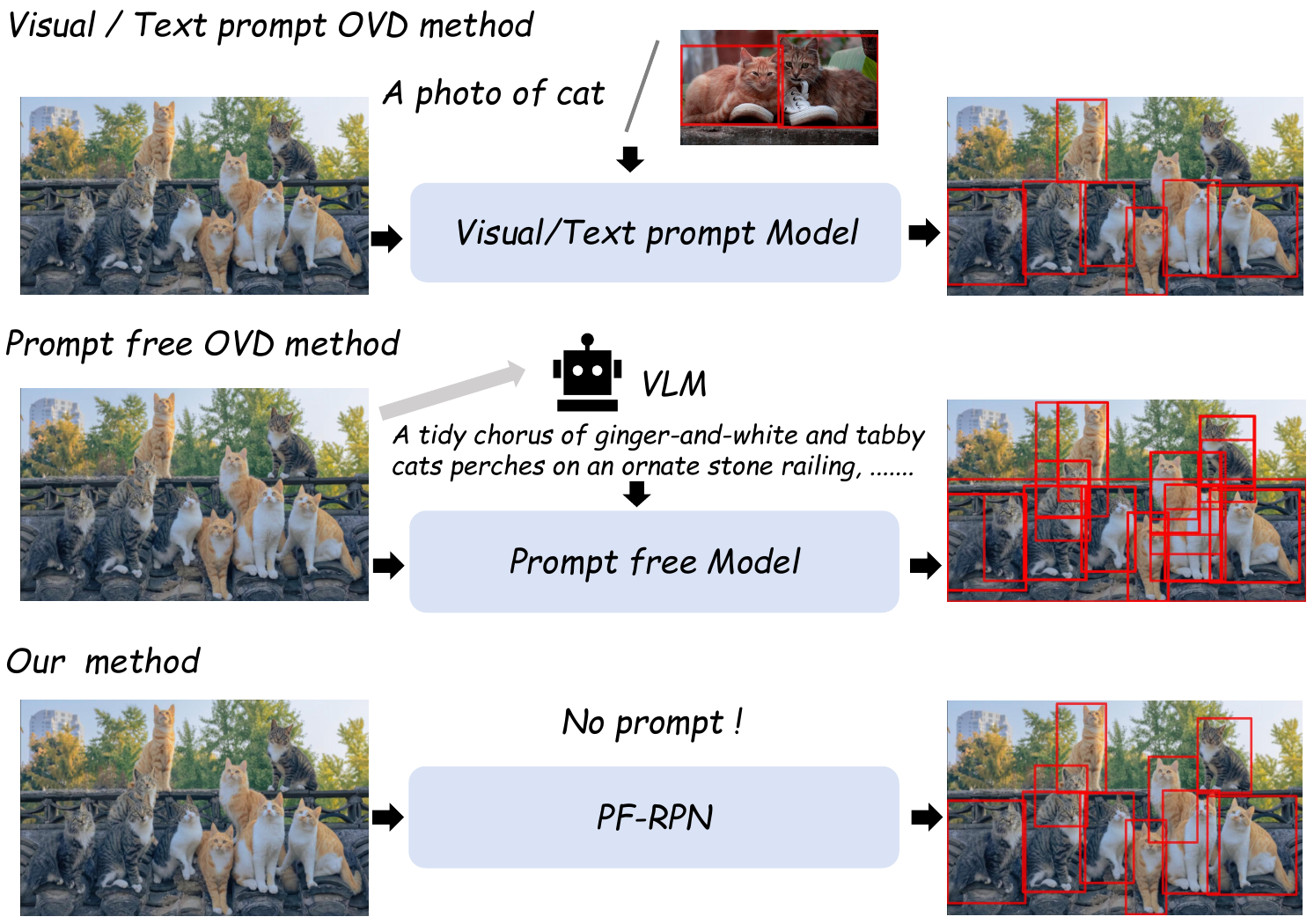}
    \vspace{-16pt}
    \caption{
      Existing visual/text prompt based OVD methods typically rely on predefined categories or exemplar images to propose potential objects for the target image. Recent prompt-free OVD methods often leverage VLMs to generate textual descriptions for the target image to localize potential objects which introduce considerable latency costs. In contrast, our PF-RPN doesn't require any external prompts and only utilizes visual features to generate high-quality proposals. Experimental results show the effectiveness of our PF-RPN in localizing potential objects.
    }        
    \label{fig:similarmap}
    \vspace{-16pt}

\end{figure}

Open-vocabulary object detection~(OVD) models~\cite{gu2021open,minderer2023scaling,Cheng2024YOLOWorld,du2022learning,wang2023learning,fu2025llmdet,kaul2023multi} have demonstrated impressive capabilities in localizing objects from unseen domains by leveraging category names or example images as prompts. 
Although OVD methods are well-suited as RPN detectors due to their strong generalization, their reliance on predefined categories and exemplar images limits flexibility in practical scenarios.
For instance, in industrial defect detection and underwater object detection scenarios, the target categories and exemplar images are often unavailable, which substantially limits the application of these models.
Although some prompt-free OVD models~\cite{lin2024generative,long2023capdet,wu2024grit,yao2024detclipv3} explore generative vision-language models (VLMs) to eliminate the need for manually provided prompts, they often introduce significant memory and latency costs.
Therefore, it is necessary to propose an efficient region proposal network that can generalize across various domains without external prompts.

In this paper, we propose a novel \textit{Prompt-Free Universal Region Proposal Network~(\ourmodel)} for localizing potential objects, which can be applied to distinct unseen domains without the need for exemplar images or textual descriptions.
Our model is optimized using limited data and can be directly applied to downstream tasks without requiring additional fine-tuning.

PF-RPN builds on the powerful OVD model by aggregating informative visual features through a learnable visual embedding, eliminating the need for manually provided prompts while retaining its strong generalization ability.
Specifically, the learnable query embedding is initialized and updated by the proposed \textit{Sparse Image-Aware Adapter~(SIA)} module, which dynamically adjusts the embedding by selectively aggregating multi-level visual features.
This adapter enables the model to capture salient visual details at various spatial resolutions, enhancing the localization of potential objects in complex visual scenes.

The SIA-adjusted learnable query embedding enables the model to identify salient objects with distinct visual appearances.
However, the embedding may still struggle to capture challenging objects with unclear visual features, such as small or occluded objects.
To mitigate this issue, we propose the \textit{Cascade Self-Prompt~(CSP)} module to identify the remaining challenging objects by iteratively refining the query embedding through a self-prompting mechanism.
The query embedding is progressively updated by aggregating multi-scale, informative visual context, enabling the model to handle ambiguities associated with small or occluded objects more effectively.

Furthermore, we observe that query embeddings near the object center tend to generate more accurate proposals than those at the object edges.
This observation motivates the design of the \textit{Centerness-Guided Query Selection~(CG-QS)} module, which selects queries based on the predicted centerness score, emphasizing the central region of objects during the query embedding selection process.
Focusing on the centermost areas helps reduce false positives and improves the quality of the proposals generated by the model.

Compared with conventional and OVD-based RPN methods, our \ourmodel~significantly improves proposal quality without requiring re-training or external prompts for unseen domains. 
Trained with limited data, \ourmodel~demonstrates strong zero-shot generalization ability, achieving consistent improvements across 19 datasets spanning diverse domains and application scenarios.
Specifically, \ourmodel~achieves 6.0/7.5/6.6 AR improvement on CD-FSOD and 4.4/5.2/5.8 AR improvement on ODinW13 with 100/300/900 candidate boxes, respectively, substantially surpassing SOTA models.
In summary, our advantages are as follows: 

\begin{itemize}
  \item 
  We propose a novel Prompt-Free Universal Region Proposal Network~(\ourmodel), a cutting-edge model which can accurately identify potential objects in practical open-world scenarios without any external prompts.
  \item 
  We propose sparse image-aware adapter, cascade self-prompting and centerness-guided query selection, enabling our model to effectively retrieve potential objects by using only the visual features.
  \item
   Our \ourmodel~achieves strong generalization performance with limited data (\eg, 5\% of COCO data) and can be directly applied to downstream tasks without additional fine-tuning. Experimental results on 19 cross-domain datasets demonstrate the effectiveness of our model.
\end{itemize}
\section{Related Works}

\textbf{Open-Vocabulary Object Detection.}
Recent progress in open-vocabulary and grounded vision–language modeling~\cite{zareian2021open,gu2021open,kuo2022f,zhong2022regionclip,zhou2022detecting,yao2023detclipv2,li2021grounded,liu2024grounding,Cheng2024YOLOWorld,wang2025yoloerealtimeseeing,du2022learning,wu2023aligning} has greatly improved detector generalization.  
GLIP~\cite{li2021grounded} unifies detection and grounding for language-aware pre-training, and Grounding DINO~\cite{liu2024grounding} enhances open-set detection via vision–language fusion. 
DetCLIPv2~\cite{yao2023detclipv2} further strengthens word–region alignment, while YOLO-World~\cite{Cheng2024YOLOWorld} and YOLOE~\cite{wang2025yoloerealtimeseeing} provide efficient vision–language fusion for accurate, real-time OVD. However, most methods still rely on text prompts or exemplar images for localization, limiting flexibility when external input is unavailable.  
Although YOLOE supports prompt-free detection, its zero-shot generalization is constrained by static text proxies.  
In contrast, our \ourmodel~learns a visual embedding and refines it through self-prompting, removing the need for text prompts while preserving strong generalization.

\textbf{Prompt-free Object Detection.}  
Recent works~\cite{lin2024generative,long2023capdet,yao2024detclipv3,wu2024grit} explore prompt-free paradigms that generate object descriptions directly.  
GenerateU~\cite{lin2024generative} formulates detection as a generative process that maps visual regions to free-form names, while CapDet~\cite{long2023capdet} bridges detection and captioning by predicting category labels or region captions.
DetCLIPv3~\cite{yao2024detclipv3} integrates a caption head into an open-set detector and leverages auto-annotated data for pre-training.  
However, such models rely on large captioners, which are computationally expensive and often biased.  
Our \ourmodel~uses a learnable embedding as a text proxy, achieving unbiased detection with low latency and memory cost.  

\textbf{Multimodal Large Language Models.}  
Multimodal Large Language Models~(MLLMs) extend LLMs with visual perception and reasoning.  
Early studies~\cite{du2022glm,li2023blip,xie2023ccmb,tong2024cambrian,xu2025llava} focused on vision-language alignment for tasks such as captioning and VQA, while later works~\cite{Qwen-VL,Qwen2-VL,Qwen2.5-VL,yao2024minicpm,yu2025minicpm,lu2024deepseek,wu2024deepseek,chen2024internvl,zhu2025internvl3,wang2025internvl3_5} (\eg, Qwen3-VL, DeepSeek-VL2) target fine-grained understanding for grounding and OCR.  
Despite their strong reasoning capability, MLLMs require massive computation and exhibit limited transfer to cross-domain detection.  
Our \ourmodel~achieves comparable zero-shot generalization without textual input or large-scale training, offering lower latency and deployment costs.

\begin{figure*}[!t]
    \centering
    \vspace{-7pt}
    \includegraphics[width=0.80\linewidth]{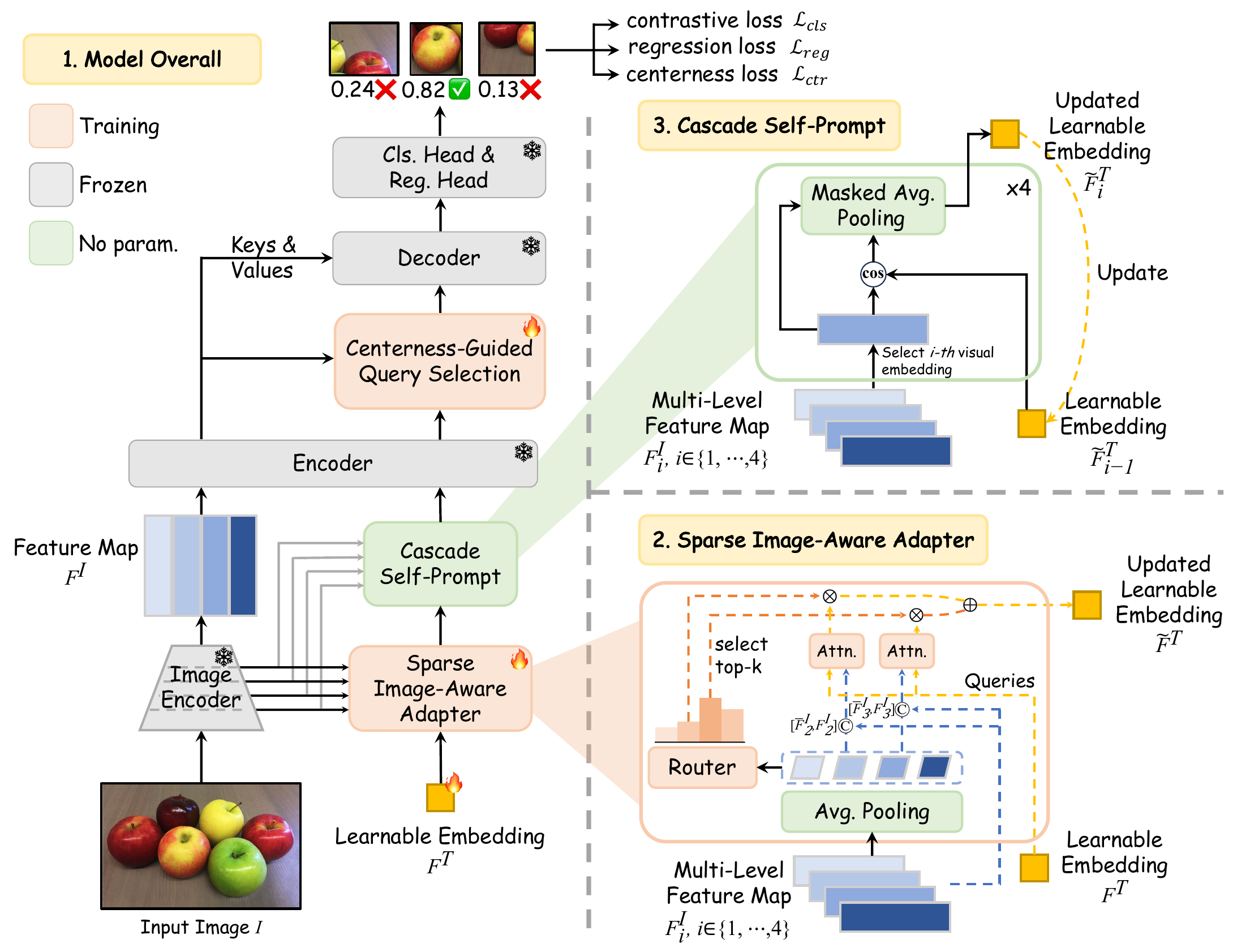}
    \vspace{-4pt}
    \caption{
    Overall architecture of our \ourmodel. It comprises three core components: (1) the Sparse Image-Aware Adapter (SIA) module, which adaptively integrates multi-level feature maps $F^I_i$ with a learnable embedding $F^T$ via a routing mechanism and cross-attention; (2) the Cascade Self-Prompt (CSP) module, which iteratively refines the embedding through masked average pooling across multiple visual levels; and (3) the Centerness-Guided Query Selection (CG-QS) module, which decodes the features into final predictions optimized by contrastive, regression, and centerness losses.
    }
    \label{fig:model}
    \vspace{-15pt}

\end{figure*}

\section{Method}
\label{sec:method}

\subsection{Method Overview}
Unlike existing prompt-free open-vocabulary object detection~(PFOVD) methods~\cite{wu2024grit,yao2024detclipv3,lin2024generative} and OVD methods~\cite{wang2025yoloerealtimeseeing,Cheng2024YOLOWorld,li2021grounded,liu2024grounding}, which rely on computationally expensive captioners to generate object names for image $I$ or require manual user input of category names or exemplar images, our \ourmodel~directly proposes potential objects across diverse domains without any text or visual prompts.

\cref{fig:model} illustrates the overall of \ourmodel. 
First, the image encoder (e.g., ResNet~\cite{he2016deep} or Swin Transformer~\cite{liu2021swin}) extracts multi-level feature maps $F^I_i \in \mathbb{R}^{H_i \times W_i \times C}, i \in \{1, \cdots, 4\}$, where $H_i \times W_i$ denotes the spatial resolution of the $i$-th feature map and $C$ is the channel dimension. 
Then, the \textit{Sparse Image-Aware Adapter~(SIA)} adaptively integrates the $k$ most informative features with the learnable embedding $F^T \in \mathbb{R}^{1 \times C}$ via a routing mechanism and cross-attention. 
Subsequently, the \textit{Cascade Self-Prompt~(CSP)} module progressively refines $F^T$ using feature maps from deep to shallow layers. 
Finally, the multi-level features $F^I_i$ are flattened into $F^I \in \mathbb{R}^{H \times W \times C}$ and used as memory, following DETR-like frameworks~\cite{carion2020end,zhu2020deformable,zhang2022dino,liu2024grounding}. 
We replace the language-guided query selection in Grounding DINO~\cite{liu2024grounding} with our \textit{Centerness-Guided Query Selection~(CG-QS)} module to decode object proposals. 
The entire framework is jointly trained on classification datasets with pseudo bounding boxes and object detection datasets.

\subsection{Sparse Image-Aware Adapter}
\label{sec:sia}

Existing OVD methods~\cite{li2021grounded,liu2024grounding,Cheng2024YOLOWorld,wang2025yoloerealtimeseeing} mainly focus on aligning image and text features for scoring detected boxes, yet they often overlook the rich multi-level visual cues from the image encoder. 
Early works~\cite{lin2017fpn,liu2018pan} reveals that feature contributions vary across levels—shallow features are beneficial for small objects, while deeper ones capture large objects—indicating that naive fusion across all levels introduces redundancy and noise. To address this, we propose the \textit{Sparse Image-Aware Adapter~(SIA)}, a Mixture-of-Experts~(MoE) module that adaptively selects and fuses the most informative feature levels with the learnable embedding $F_T$. 
Inspired by visual feature-based prompt tuning~\cite{zhou2022conditional}, SIA replaces text embeddings in pretrained OVDs~(\eg, Grounding DINO~\cite{liu2024grounding}) with image-derived representations, bridging the modality gap.

Given the multi-level feature maps $F_I^i$, a global average pooling layer extracts compact features $\bar{F}_I^i \in \mathbb{R}^C$. 
An MoE router predicts their importance $w_i = \textit{Router}(\bar{F}_I^i)$, where $\textit{Router}$ is a lightweight MLP. 
We then select the top-$k$~($k \le 4$) feature levels and normalize their weights via softmax. 
Finally, $F_T$ acts as the query and the concatenated features $\left[\bar{F}_{\sigma(j)}^I, F_{\sigma(j)}^I\right]$ serve as key–value pairs in cross-attention~\cite{vaswani2017attention,rao2022denseclip} to produce the updated embedding:
\begin{equation}
    \Tilde{F}^T = \sum_{j=1}^k \Tilde{w}_{\sigma(j)} \cdot \text{Attn}\left(F_T, [\bar{F}_{\sigma(j)}^I, F_{\sigma(j)}^I]\right),
\end{equation}
where $\sigma(j)$, $1 \le j \le k$ denotes the selected feature levels.

The proposed SIA module sparsely adapts multi-level visual features to the learnable embedding while maintaining consistency between object scales and feature levels. 
Moreover, by leveraging both global features $\bar{F}_{\sigma(j)}^I$ and local features $F_{\sigma(j)}^I$, the learnable embedding is enriched with both coarse- and fine-grained visual cues. 
As illustrated in \cref{fig:moe}, SIA significantly enhances the localization capability of the learnable embedding by emphasizing semantically relevant object regions and suppressing background noise. 
However, background activations are still observed, suggesting that a single-step adaptation is insufficient.
To further refine the embedding and achieve more precise localization, we introduce the CSP module in the next section.

\subsection{Cascade Self-Prompt}
\label{sec:csp}

While the SIA module enriches the learnable embedding $\Tilde{F}^T$ with scale-relevant cues and enhances its localization ability, we observe that some background regions may still be partially activated as shown in \cref{fig:moe}.
This suggests that a single-step adaptation remains insufficient to fully suppress noisy responses. 
To further purify the embedding, we design a refinement mechanism that leverages the embedding’s own visual activations.

Empirically, object-internal features exhibit stronger localization ability than the learnable embedding itself, and this finding is proved in our supplementary materials.

This motivates an iterative refinement scheme in which activated visual feature progressively guide $\Tilde{F}^T$ toward more discriminative representations. 
Moreover, since deeper layers encode high-level semantics while shallower layers capture fine-grained structural details~\cite{zeiler2014visualizing,lin2017feature}, we perform the refinement in a deep-to-shallow cascade—first aggregating semantics, then integrating structure.

Based on these insights, we propose the \textit{Cascade Self-Prompt~(CSP)} module, which iteratively refines $\Tilde{F}^T$ using multi-level features $F_i^I$. 
Starting from $\Tilde{F}_0^T = \Tilde{F}^T$, we generate a similarity mask at each level:
\begin{equation}
    M_i = \mathbbm{1}\left(\cos\left(\Tilde{F}_{i-1}^T, F_i^I\right) > \delta\right),
    \label{eq:mask}
\end{equation}
where $\delta$ is a manually set threshold (set to 0.3), $\cos$ denotes the cosine similarity, and $\mathbbm{1}$ is the indicator function.
The embedding is then updated via masked average pooling:
\begin{equation}
    \Tilde{F}_i^T = \Tilde{F}_{i-1}^T + \textit{MAP}(M_i, F_i^I),
\end{equation}
where $\textit{MAP}$ denotes the masked average pooling.
By cascading this process from deep to shallow layers, CSP progressively expands object-consistent activations while suppressing background noise. 
Guided by the strong prior from SIA, the refinement jointly optimizes visual consistency and scoring reliability, yielding more precise and robust localization.
\cref{fig:cascade} illustrates the effectiveness of this iterative process.
To achieve an optimal balance between accuracy and efficiency, we set the number of iteration to $3$.

\begin{table*}[!t]
\centering
\vspace{-7pt}
\caption{
Comparison with OVD models, MLLMs, and RPNs. 
The best results are highlighted in \textbf{bold}. 
$AR_{100/300/900}$ denotes the average recall under 100/300/900 candidate boxes, respectively, and $AR_{s/m/l}$ denotes the average recall for small/medium/large objects. 
$\dagger$~indicates using the original class names from the corresponding dataset as text input, and $\ddagger$~indicates replacing the original class names with ``object'' to obtain a prompt-free setting. 
Our method achieves the best performance on both the CD-FSOD and ODinW13 benchmarks.
}
\label{tab:results}
\vspace{-7pt}
\small
\renewcommand{\arraystretch}{1.0} 
\begin{tabular}{@{\hspace{4pt}}ccc*{6}{c}@{\hspace{4pt}}}
\toprule
\rowcolor{HeaderGray}
\multirow{1}{*}{Datasets} & 
\multirow{1}{*}{Methods} & 
\multirow{1}{*}{Prompt Free} & 
\multirow{1}{*}{$AR_{100}$} & 
\multirow{1}{*}{$AR_{300}$} & 
\multirow{1}{*}{$AR_{900}$} & 
\multirow{1}{*}{$AR_{s}$} & 
\multirow{1}{*}{$AR_{m}$} & 
\multirow{1}{*}{$AR_{l}$} \\
\midrule
& GDINO$^\dagger$~\cite{liu2024grounding} & \textcolor{Red}{\ding{55}} &52.9&53.5& 54.7&31.1&41.6&63.9\\
\rowcolor{LightBlueRow} \cellcolor{white}
& GDINO$^\ddagger$~\cite{liu2024grounding} & \textcolor{Green}{\ding{51}} &54.7&57.8& 61.6&34.1&49.3&67.0 \\
& YOLOE-v8-L$^\dagger$~\cite{wang2025yoloerealtimeseeing} & \textcolor{Red}{\ding{55}} &44.4&46.2&47.1&21.6&36.6&54.9\\
\rowcolor{LightBlueRow} \cellcolor{white}
& YWorldv8-L$^\dagger$~\cite{Cheng2024YOLOWorld} & \textcolor{Red}{\ding{55}} & 49.6&	51.1&51.6&25.1&	42.7&60.6\\
& Qwen-VL$^\dagger$~\cite{Qwen2.5-VL} & \textcolor{Red}{\ding{55}} & 20.1 & 20.1 & 20.1 & 1.0 & 3.0 &26.5\\
\rowcolor{LightBlueRow} \cellcolor{white}
&GLIP$^\dagger$~\cite{li2021grounded}&\textcolor{Red}{\ding{55}}& 47.6& 47.6& 	47.6 & 	21.2 & 	34.6 &	56.0\\
& GenerateU~\cite{lin2024generative} &\textcolor{Green}{\ding{51}}& 47.7 & 54.1 & 55.7 & 28.1 & 48.3 & 69.4\\
\rowcolor{LightBlueRow} \cellcolor{white}
& Open-Det~\cite{caoopen} &\textcolor{Green}{\ding{51}}& 36.6 & 46.3 & 54.3 & 28.2 & 45.3 & 67.7 \\
&RPN~\cite{ren2016faster}&\textcolor{Green}{\ding{51}}&32.0&39.0&	45.7&	29.9&	43.0&	54.3\\
\rowcolor{LightBlueRow} \cellcolor{white}
&Cascade RPN~\cite{vu2019cascade}&\textcolor{Green}{\ding{51}}&45.8&52.0&56.9&31.1&50.5&66.0\\
\cmidrule{2-9}
\rowcolor{OursHighlight} \cellcolor{white} \multirow{-12}{*}{\rotatebox[origin=c]{90}{\textbf{CD-FSOD}}}
& \textbf{Ours} & \textcolor{Green}{\ding{51}} & \textbf{60.7} & \textbf{65.3} & \textbf{68.2} & \textbf{38.5} & \textbf{61.9} & \textbf{80.3}\\
\midrule
& GDINO$^\dagger$~\cite{liu2024grounding} & \textcolor{Red}{\ding{55}} & 72.1 & 73.4 & 74.0 & \textbf{45.6} & 61.7 & 79.2\\
\rowcolor{LightBlueRow} \cellcolor{white}
& GDINO$^\ddagger$~\cite{liu2024grounding} & \textcolor{Green}{\ding{51}} &69.1&70.9& 72.4&40.8&64.6&78.4\\
& YOLOE-v8-L$^\dagger$~\cite{wang2025yoloerealtimeseeing} & \textcolor{Red}{\ding{55}} & 66.6 & 67.8 & 68.3 & 39.2 & 57.8 & 72.8\\
\rowcolor{LightBlueRow} \cellcolor{white}
& YWorldv8-L$^\dagger$~\cite{Cheng2024YOLOWorld} & \textcolor{Red}{\ding{55}} & 69.1&	70.3&71.5&37.5&62.2&75.4\\
&GLIP$^\dagger$~\cite{li2021grounded}&\textcolor{Red}{\ding{55}}& 69.8& 	69.8& 	69.8 & 	33.2 & 	50.9 &	75.2\\
\rowcolor{LightBlueRow} \cellcolor{white}
& GenerateU~\cite{lin2024generative}  &\textcolor{Green}{\ding{51}}& 67.3 & 71.5 & 72.2 & 32.8 & 63.1 & 80.0\\
& Open-Det~\cite{caoopen}  &\textcolor{Green}{\ding{51}}& 53.9 & 62.9 & 69.1 & 27.7 & 59.8 & 76.6\\
\rowcolor{LightBlueRow} \cellcolor{white}
&RPN~\cite{ren2016faster}&\textcolor{Green}{\ding{51}}&49.0&	52.4&55.7&	35.3&54.0&59.8\\
&Cascade RPN~\cite{vu2019cascade}&\textcolor{Green}{\ding{51}}&60.9&65.5&70.2&40.3&65.5&75.0\\
\cmidrule{2-9}
\rowcolor{OursHighlight} \cellcolor{white} \multirow{-11}{*}{\rotatebox[origin=c]{90}{\textbf{ODinW13}}}
& \textbf{Ours} & \textcolor{Green}{\ding{51}} & \textbf{76.5} & \textbf{78.6} & \textbf{79.8} & 45.4 & \textbf{71.9} & \textbf{85.8}\\
\bottomrule
\end{tabular}
\vspace{-7pt}
\end{table*}

\subsection{Centerness-Guided Query Selection}
After CSP module, we can localize potential object regions and obtain their corresponding queries. However, the importance of each query largely depends on its spatial location.
As shown in \cref{fig:center}, queries located near the object center tend to produce more accurate proposals than those near object boundaries.
Therefore, we propose the \textit{Centerness-Guided Query Selection (CG-QS)} module to estimate the likelihood that each query lies near the object center.

Specifically, a lightweight MLP is employed as a center scoring network to generate a center score $g_i$ for each query $f_i$. Meanwhile, we compute the distances from the query to the left, right, top, and bottom edges of the corresponding ground-truth box to derive the center supervision $c_i$:
\begin{equation}
    c_i = \sqrt{ \frac{\min(l, r)}{\max(l, r)} \times \frac{\min(t, b)}{\max(t, b)} }.
\end{equation}  

When a query is closer to the ground-truth box center, the corresponding supervision $c_i$ approaches 1, and the network is trained to make the predicted score $g_i$ match $c_i$. The centerness loss is then defined as the L1 distance between the predicted center score $g_i$ and its supervision $c_i$, $\mathcal{L}_{\textit{ctr}} = \sum_{i=1}^{N} \| g_i - c_i \|_1$, where $N$ denotes the total number of queries and $\|\cdot\|_1$ represents the L1 loss.

The proposed CG-QS module effectively prioritizes visual embeddings near object centers. During both training and inference, given classification scores computed by the dot product between the learnable embedding and the queries, we combine the center scores generated by the scoring network with these classification scores for query selection, and then use the resulting scores to determine the final candidate query set.

\subsection{Objective Loss}

Previous work~\cite{fan2022few} shows that the fine-tuning stage of detectors introduces bias into the image encoder, since detection models are fine-tuned on detection datasets, whereas the image encoder is pretrained on classification datasets, \eg, ImageNet~\cite{deng2009imagenet}.
To alleviate this bias, we jointly fine-tune our \ourmodel~on 5\% of the data from ImageNet with pseudo bounding boxes and COCO~\cite{lin2014microsoft}, thereby reducing the distribution gap between classification and detection data.

Following DETR-like frameworks~\cite{carion2020end,zhu2020deformable,zhang2022dino,liu2024grounding,li2021grounded}, we employ the L1 loss and the GIoU loss~\cite{rezatofighi2019generalized} as the regression loss $\mathcal{L}_{\textit{reg}}$, and use a contrastive loss between queries and the learnable embedding $\Tilde{F}_4^T$ for classification scoring.

To prevent a few experts from being over-activated while others remain rarely used—resulting in load imbalance—we introduce an auxiliary loss $\mathcal{L}_{\textit{rt}} = \text{std}(w_i),  i \in \{1, \cdots, 4\}$ on the expert weights $w_i$ to balance the load across experts and fully exploit the multi-level feature maps, where $\text{std}$ denotes the empirical standard deviation. Minimizing $\mathcal{L}_{\textit{rt}}$ encourages the expert weights $w_i$ from the router to be more evenly distributed, improving load balance.
Finally, the overall objective function is formulated as:
\begin{equation}
    \mathcal{L} = \mathcal{L}_{\textit{reg}} + \mathcal{L}_{\textit{cls}} + \mathcal{L}_{\textit{rt}} + \lambda \mathcal{L}_{\textit{ctr}},
\end{equation}
where $\mathcal{L}_{\textit{reg}}$ and $\mathcal{L}_{\textit{cls}}$ follow the same configurations as in Grounding DINO~\cite{liu2024grounding}.

\section{Experiments}

We adopt Grounding DINO~\cite{liu2024grounding} with a Swin-B backbone as our baseline.
Our model is trained on $5\%$ of the COCO~\cite{lin2014microsoft} dataset (80 classes) and $5\%$ of the ImageNet~\cite{deng2009imagenet} dataset (1000 classes) and can be directly applied to downstream tasks without any further fine-tuning.
Following previous work~\cite{li2021grounded}, we evaluate our model on the ODinW13 benchmark, which includes datasets from diverse domains such as wildlife photography, household objects, and aerial imagery.
To further assess the generalization of our model, we also evaluate our model on the CD-FSOD benchmark, which consists of six cross-domain datasets with distinct domain shifts: ArTaxOr~\cite{mazen2023arthropod} (insect images), Clipart1k~\cite{inoue2018cross} (hand-drawn cartoon images), DIOR~\cite{li2020object} (remote sensing images), DeepFish~\cite{saleh2020realistic} (underwater fish images), NEU-DET~\cite{huang2020surface} (industrial defect images), and UODD~\cite{wei2024underwater} (marine organism images). In our experiments, we use Average Recall (AR) as the evaluation metric to evaluate our \ourmodel's ability to propose potential objects. All experiments are conducted on four NVIDIA RTX 4090 GPUs. 
\subsection{Quantitative Results}

\noindent{\bf Comparison with OVD Models, RPNs and MLLMs.}
As shown in Table~\ref{tab:results}, we compare our \ourmodel~with typical open-vocabulary object detection (OVD) models. For OVD models, we feed the class names from the corresponding dataset into the model to obtain detection boxes that serve as proposals. Meanwhile, to further investigate the impact of text prompts on model performance, we also evaluate its performance under the prompt-free setting by replacing the class names with ``object'' as the model text input. Our \ourmodel~outperforms the baseline model Grounding DINO, achieving improvements of 7.8/11.8/13.5 AR on the CD-FSOD benchmark under 100/300/900 candidate boxes, respectively. On the ODinW13 benchmark, our \ourmodel~further surpasses Grounding DINO by 4.4/5.2/5.8 AR under 100/300/900 candidate boxes. Compared with the OVD model YOLOE~\cite{wang2025yoloerealtimeseeing}, our \ourmodel~achieves performance gains of 16.3/19.1/21.1 AR. To further assess the generalization of our \ourmodel, we also compare it with MLLMs. Specifically, compared with Qwen2.5-VL-7B~\cite{Qwen2.5-VL}, our \ourmodel~obtains improvements of 40.6/45.2/48.1 AR under 100/300/900 candidate boxes. In addition, compared with the Cascade RPN~\cite{vu2019cascade}, our \ourmodel~improves performance by 15.6/13.1/9.6 AR on the ODinW13 benchmark.

\begin{figure*}[!t]
    \centering
    \includegraphics[width=0.8\linewidth]{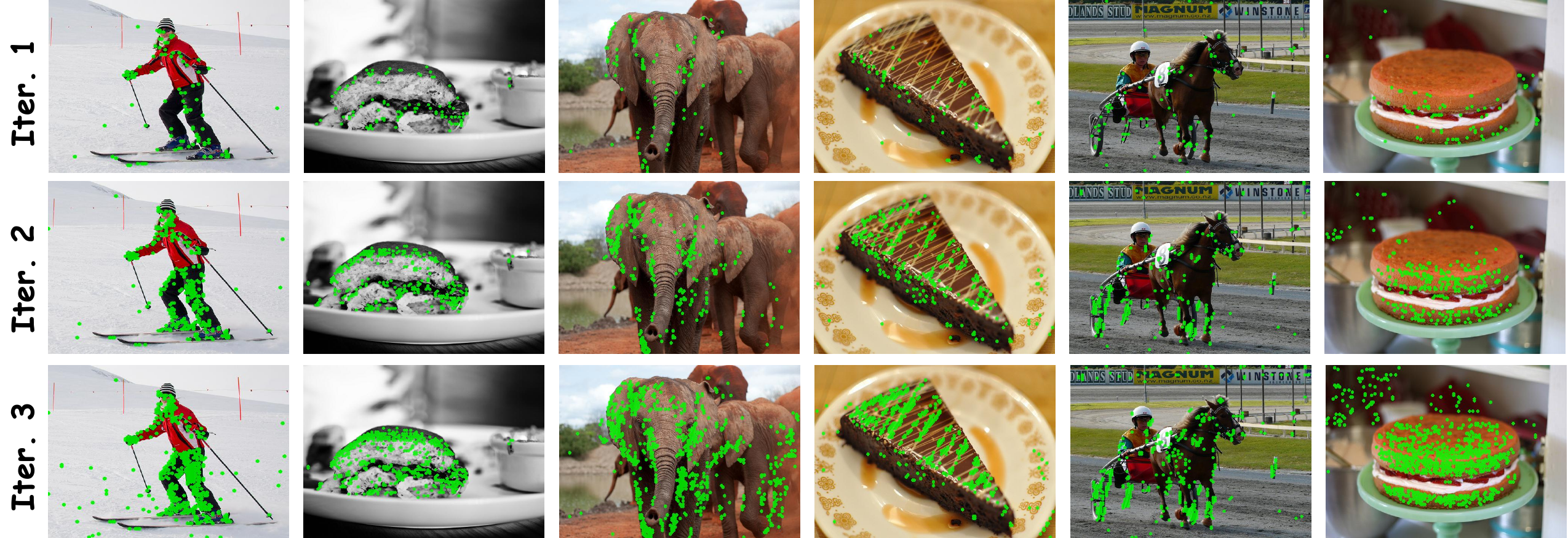}
    \vspace{-4pt}
    \caption{
        \textbf{Effect of iterations in the Cascade Self-Prompt module.}
        Visualization of region selection across different Cascade Self-Prompt iterations. Green points indicate the object regions selected by the model in the current iteration. As the number of iterations increases, the model progressively selects more object regions in the image, demonstrating the effectiveness of our cascade self-prompt mechanism.
    }        
    \label{fig:cascade}
    \vspace{-6pt}
\end{figure*}
\begin{figure*}[!t]
    \centering
    \includegraphics[width=0.8\linewidth]{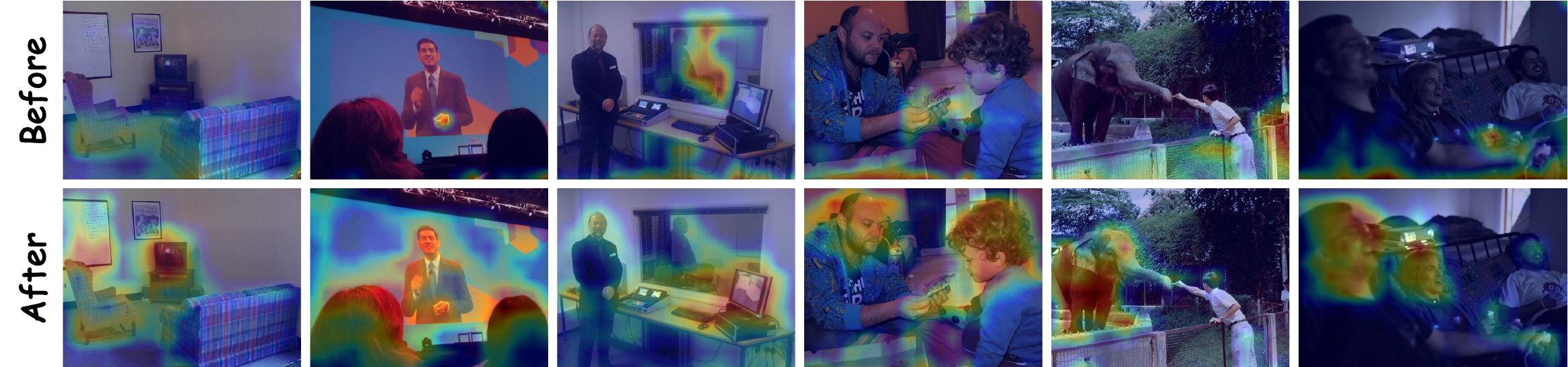}
    \vspace{-4pt}
    \caption{
    \textbf{Effect of the Sparse Image-Aware Adapter.}
    Visualization of similarity heatmaps between the learnable embedding and image features before and after the module update.
    Each pair of heatmaps (top: before, bottom: after) corresponds to the same image.
    After the update, the learnable embedding exhibits stronger responses in semantically relevant regions, indicating improved alignment between visual and learned representations and providing a stronger prior for the cascade self-prompt module.
    }\label{fig:moe}
    \vspace{-6pt}
\end{figure*}

\noindent{\bf Module Ablation Studies.}
To evaluate the contribution of each module, we conduct the module ablation study on the CD-FSOD benchmark. As shown in \cref{tab:ablation}, adding the SIA module raises the average performance to 57.8 $AR_{100}$, outperforming the baseline and indicating that visual features are more effective than text for localizing potential objects. Building on this, adding both the SIA and CSP modules further improves the performance to 60.2 $AR_{100}$, showing that the cascaded self-prompt strategy effectively reduces missed detections by iteratively updating the learnable embedding to retrieve more potential objects. Adding the SIA module and CG-QS modules improves performance to 59.6 $AR_{100}$, demonstrating that the center scoring network can accurately assess proposal quality and help the model select high-quality proposals. When combining all modules, our approach achieves the best performance of 60.7 $AR_{100}$, confirming the complementarity among these modules.

\begin{table}[!t]
\centering
\caption{Results of module ablation studies. SIA denotes the Sparse Image-Aware Adapter, CSP denotes the Cascade Self-Prompt module and CG-QS denotes the Centerness-Guided Query Selection module. The best results are highlighted in \textbf{bold}.}
\label{tab:ablation}
\vspace{-8pt}
\small
\setlength{\tabcolsep}{2pt}
\begin{tabular}{@{\hspace{2pt}}ccccccccccc@{\hspace{2pt}}}
\toprule
\rowcolor{HeaderGray}
{SIA} & {CSP} & {CG-QS} & {$AR_{100}$} & {$AR_{300}$} & {$AR_{900}$} & {$AR_{s}$} & {$AR_{m}$} & {$AR_{l}$}\\
\midrule

$\times$ & $\times$ & $\times$  &52.9&53.5& 54.7&31.1&41.6&63.9\\
\rowcolor{LightBlueRow}
$\checkmark$ &  $\times$ &  $\times$ & 57.8&	63.1&	66.7&	47.9&	58.2&	77.5\\
$\times$ & $\checkmark$ & $\times$   & 58.1&	62.9&	65.8&	37.7&	56.7&	77.0\\
\rowcolor{LightBlueRow}
$\times$ & $\times$ & $\checkmark$  & 54.4&	56.9&	60.2&	21.9&	50.6&	79.8\\
$\checkmark$ & $\times$ & $\checkmark$  & 60.2&	64.7&	68.0&	47.1&	60.7&	79.1\\
\rowcolor{LightBlueRow}
$\times$ & $\checkmark$ & $\checkmark$ &  59.6	&64.3&	67.3&	45.2&	60.6&	78.1\\
$\checkmark$ & $\checkmark$ & $\times$ & 60.2&	65.2&	67.7&	43.4&	60.0& 	78.4\\
\rowcolor{LightBlueRow}
$\checkmark$ & $\checkmark$ & $\checkmark$  & \textbf{60.7}&	\textbf{65.3}&	\textbf{68.2}&	\textbf{38.5}&	\textbf{61.9}&	\textbf{80.3}\\

\bottomrule
\end{tabular}
\vspace{-12pt}
\end{table}

\noindent{\bf Data Ablation Studies.}
To investigate the influence of training data scale on our model, we conduct a data ablation experiment on the CD-FSOD benchmark. 
As shown in \cref{tab:dataala}, increasing the proportion of detection data from COCO leads to consistent improvements in average recall (AR). 
Notably, the performance gain from using 1\% to 5\% of COCO is significantly larger than that from 5\% to 10\%, indicating diminishing returns when further expanding the data scale. 
Therefore, we adopt 5\% of COCO as a trade-off between performance and efficiency. Furthermore, introducing classification data (ImageNet) yields an additional improvement in AR, demonstrating its effectiveness in alleviating the bias in the image encoder caused by detection-only training. 
This confirms that a small amount of classification data helps enhance cross-modal alignment and improves the generalization ability of our model.

\begin{table}[!t]
\centering
\caption{Results of data ablation studies. COCO denotes the COCO dataset and IN denotes the ImageNet dataset, The best results are highlighted in \textbf{bold}.}
\vspace{-8pt}
\label{tab:dataala}
\small
\setlength{\tabcolsep}{4pt}
\begin{tabular}{@{\hspace{2pt}}cccccccc@{\hspace{2pt}}}
\toprule
\rowcolor{HeaderGray}
COCO & IN & {$AR_{100}$} & {$AR_{300}$} & {$AR_{900}$} & {$AR_{s}$} & {$AR_{m}$} & {$AR_{l}$}\\
\midrule
$1\%$  & 0      & 57.2 & 63.7 & 68.0  & \textbf{42.2} & 60.9 & 76.6\\
\rowcolor{LightBlueRow}
$5\%$  & 0      & 58.8 & 64.1 & 67.8 & 40.2        & 59.0  &78.6\\
$10\%$ & 0      & 59.0 & 64.1 & 67.5 & 38.9 & 59.1&78.9\\
\rowcolor{LightBlueRow}
$1\%$  & $5\%$  & 59.1 & 63.3 & 66.8 & 36.2 & 59.9&79.8\\
$5\%$  & $5\%$  & 60.7 & 65.3 &	68.2 & 38.5 &	\textbf{61.9}&	80.3\\
\rowcolor{LightBlueRow}
$10\%$ & $5\%$  & \textbf{60.9}&\textbf{65.5}&\textbf{68.5}&39.6&61.6&\textbf{80.5}\\
\bottomrule
\end{tabular}
\vspace{-4pt}
\footnotesize 
\end{table}

\noindent{\bf Comparison with Different Backbone.}
As shown in \cref{tab:transfer}, the experimental results demonstrate that our model achieves strong performance across different backbone. Specifically, when integrating our model with a ResNet-50 backbone~\cite{he2016deep}, the performance improves by 5.2 $AR_{100}$, while using a Swin-B backbone~\cite{liu2021swin} leads to an improvement of 7.8 $AR_{100}$.

\begin{table}[!t]
\centering
\caption{Results with different backbones. The best results for each baseline are highlighted in \textbf{bold}. Our model effectively improves the performance of detectors with different backbone.}
\label{tab:transfer}
\vspace{-8pt}
\small
\setlength{\tabcolsep}{2pt}
\begin{tabular}{@{\hspace{2pt}}ccccccccccc@{\hspace{2pt}}}
\toprule
\rowcolor{HeaderGray}
{Method} & {$AR_{100}$} & {$AR_{300}$} & {$AR_{900}$} & {$AR_{s}$} & {$AR_{m}$} & {$AR_{l}$}\\
\midrule
GDINO (SwinB) &52.9&53.5& 54.7&31.1&41.6&63.9\\
\rowcolor{OursHighlight} 
~~~~~~~~~~+ ours & \textbf{60.7} & \textbf{65.3} & \textbf{68.2} & \textbf{38.5} & \textbf{61.9} & \textbf{80.3}\\
\midrule
GDINO (ResNet50) &40.4&	43.4&	44.6&	21.7&	32.5&	54.9\\
\rowcolor{OursHighlight}
~~~~~~~~~~ + ours  & \textbf{45.6} & \textbf{51.4} & \textbf{58.0} & \textbf{28.1} & \textbf{50.6} & \textbf{73.7} \\

\bottomrule
\end{tabular}
\vspace{-0.2in}
\footnotesize 
\end{table}

\noindent\textbf{Ablation Study on MoE module in the SIA.} To verify the effectiveness of the MoE module, we conduct ablation experiments as shown in \cref{tab:moe}. The removal of MoE leads to a consistent performance drop, further validating its importance.
The experimental results demonstrate that while the attention mechanism can suppress irrelevant information, it operates \textit{only within} individual feature levels and is \textit{unable to select across levels}. In contrast, the MoE module \textit{filters out irrelevant feature levels before} the attention stage. Since objects of different scales are best represented at distinct feature levels, relying solely on attention would inevitably introduce noise from non-informative levels.

\begin{table}[!t]
\centering
\vspace{-0.05in}
\caption{Ablation study on the effectiveness of the MoE module. The best results for each baseline are highlighted in \textbf{bold}.}
\label{tab:moe}
\vspace{-8pt}
\resizebox{\linewidth}{!}{
    \footnotesize
    \setlength{\tabcolsep}{4pt}
    \renewcommand{\arraystretch}{1.1}
    \begin{tabular}{@{}c c c c c c c c c@{}}
    \toprule
    \rowcolor{HeaderGray}
    Dataset & Method & $AR_{100}$ & $AR_{300}$ & $AR_{900}$ & $AR_{s}$ & $AR_{m}$ & $AR_{l}$ \\
    \midrule
    & w/o MoE & 58.6 & 61.2 & 63.5 & 37.1 & 56.5 & 73.5 \\
    \rowcolor{OursHighlight} \cellcolor{white} \multirow{-2}{*}{\textbf{CD-FSOD}}
    & \textbf{Ours} & \textbf{60.7} & \textbf{65.3} & \textbf{68.2} & \textbf{38.5} & \textbf{61.9} & \textbf{80.3} \\
    \midrule
    & w/o MoE & 68.7 & 71.7 & 73.8 & 37.9 & 67.0 & 80.7 \\
    \rowcolor{OursHighlight} \cellcolor{white} \multirow{-2}{*}{\textbf{ODinW13}}
    & \textbf{Ours} & \textbf{76.5} & \textbf{78.6} & \textbf{79.8} & \textbf{45.4} & \textbf{71.9} & \textbf{85.8}\\
    \bottomrule
    \end{tabular}
}
\vspace{-0.15in}
\end{table}

\noindent{\bf Integrating into Well-Train Detectors.} 
To evaluate the extensibility of our model, we integrate our model into existing well-trained RPN-based detectors. As shown in \cref{tab:devit}, when we replace the original RPN in DE-ViT~\cite{xu2023devit} with our proposed module, the detector achieves an improvement of 3.7 AP on the COCO dataset. Furthermore, to evaluate the generalization of our model in cross-domain scenarios, we integrate our model into the cross-domain detector CD-ViTO~\cite{fu2024cross} and evaluate its performance on the CD-FSOD benchmark following the original setting. Experiment result shows that Integrating our model results in an improvement of 5.5 AP on the CD-FSOD benchmark.

\begin{table}[!t]
\vspace{-0.05in}
\centering
\caption{Results of integrating our model into well-trained detectors. OD denotes the conventional object detection task evaluated on the COCO dataset and CDFSOD denotes the cross-domain few-shot object detection task evaluated on the CD-FSOD benchmark. The best results are highlighted in \textbf{bold}.
}
\label{tab:devit}
\small
\setlength{\tabcolsep}{3pt}
\vspace{-8pt}
\begin{tabular}{@{}cccccccc@{}}
\toprule
\rowcolor{HeaderGray}
Method & Task & $AP$ & $AP_{50}$ & $AR_{75}$ & $AP_{s}$ & $AP_{m}$ & $AP_{l}$\\
\midrule
DE-VIT & OD     & 33.0 & 52.9 & 33.7 & 17.1 & 35.4 & 47.3\\
\rowcolor{OursHighlight}
+ ours & OD    & \textbf{36.7} & \textbf{56.2} & \textbf{38.8} & \textbf{23.5} & \textbf{40.1} & \textbf{50.7}\\
\midrule
CD-VITO & CD-FSOD & 29.6 & 49.1 & 30.0 & 5.7  & \textbf{24.4} & 37.6\\
\rowcolor{OursHighlight}
+ ours  & CD-FSOD & \textbf{35.1} & \textbf{59.6} & \textbf{35.0} & \textbf{9.0} & 22.6 & \textbf{41.9}\\
\bottomrule
\end{tabular}
\vspace{-0.15in}
\footnotesize
\end{table}

\begin{figure*}[!t]
    \centering
    \vspace{-4pt}
    \includegraphics[width=0.78\linewidth]{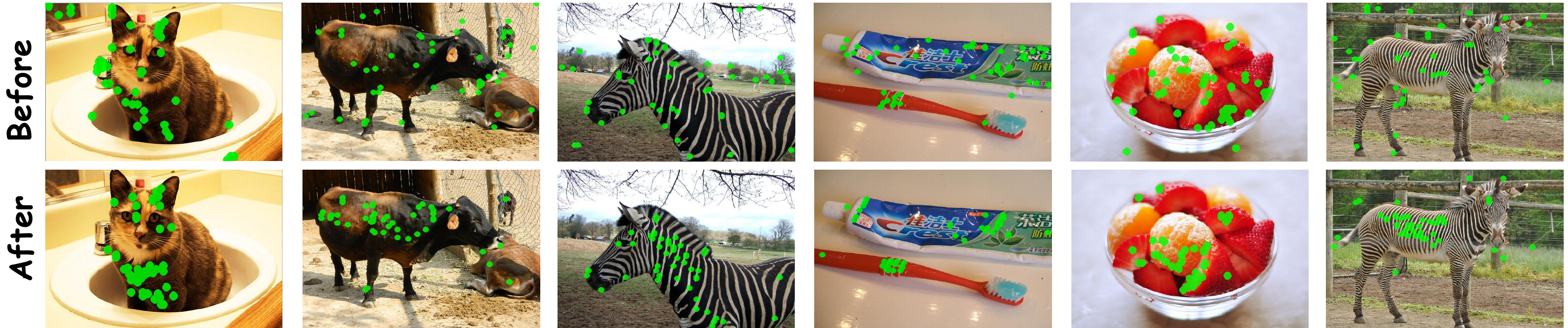}
    \vspace{-4pt}
    \caption{
        \textbf{Effect of the Centerness-Guided Query Selection.}
        Visualization of query selection before and after applying the Centerness-Guided Query Selection (CG-QS) module.
        Each pair of heatmaps (top: before, bottom: after) corresponds to the same image. After applying the CG-QS module, the model tends to select queries near object centers, thereby generating more accurate proposals.
    }        
    \label{fig:cencervisual}
    \vspace{-4pt}
\end{figure*}
\begin{figure*}[!t]
    \centering
    \includegraphics[width=0.75\linewidth]{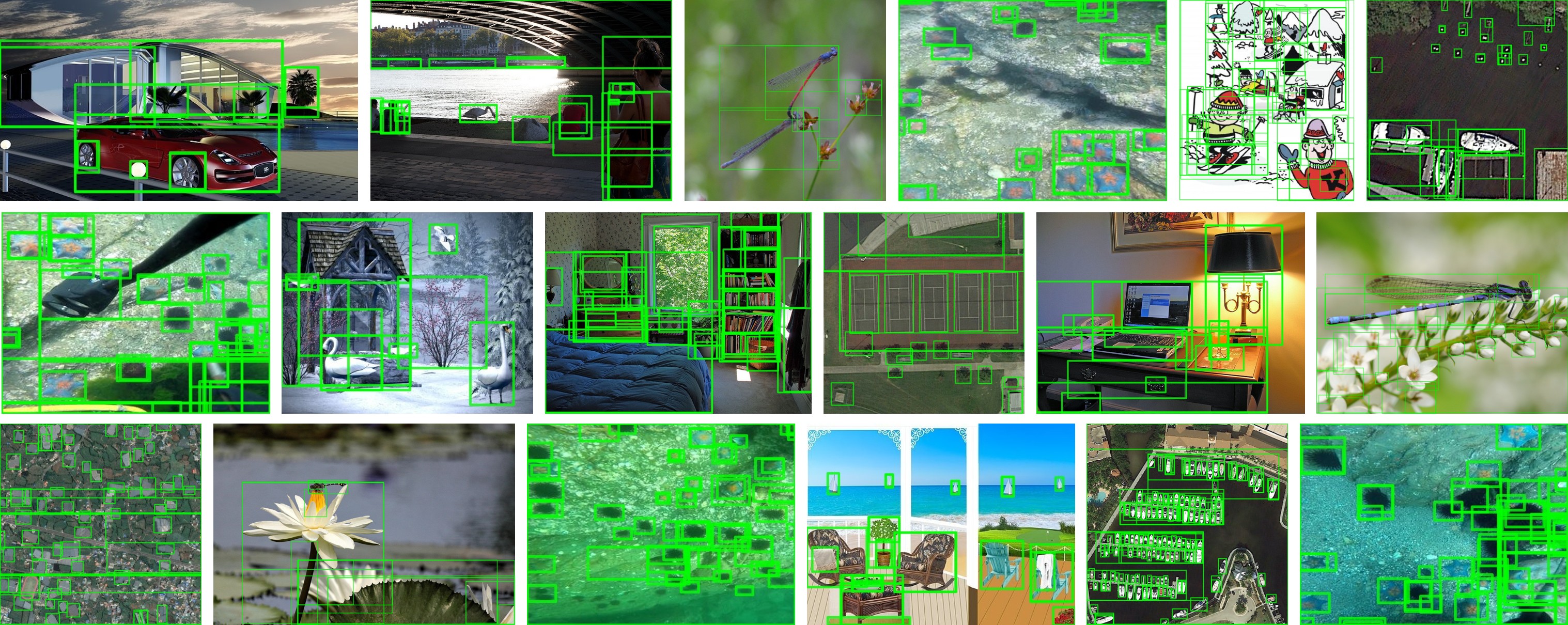}
    \vspace{-4pt}
    \caption{
    Qualitative results of \textbf{\ourmodel}~on several object detection benchmarks.
    \ourmodel~exhibits strong cross-domain generalization and localization ability, accurately proposing potential object regions without any domain-specific fine-tuning.
    }     
    \label{fig:demo}
    \vspace{-8pt}
\end{figure*}

\begin{figure}[!t]
    \centering
    \includegraphics[width=0.8\linewidth]{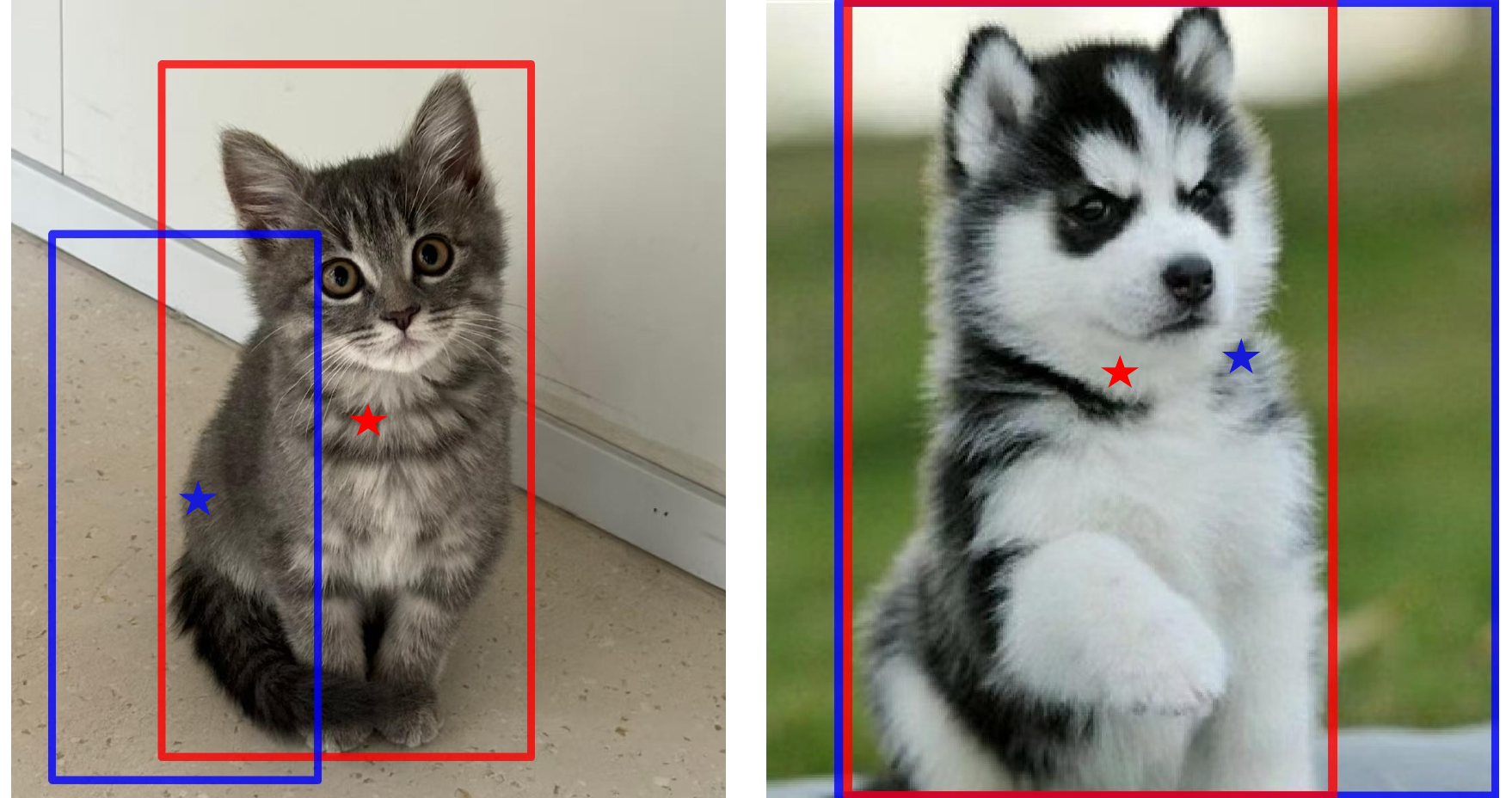}
    \vspace{-7pt}
    \caption{
    Prediction box comparison. The red box corresponds to the query indicated by the red star, while the blue box corresponds to the query indicated by the blue star. It can be observed that the query located near the object center produces a more accurate bounding box than queries near object boundaries.
    }        
    \label{fig:center}
    \vspace{-10pt}
\end{figure}

\subsection{Qualitative Visualizations}
\noindent{\bf Cascade Self-Prompt Module.} In the Cascade Self-Prompt (CSP) module, our key idea is to use visual features from some objects to iteratively retrieve the remaining potential objects. To validate the effectiveness of this paradigm, we visualize the selected image regions at different iterations. As shown in \cref{fig:cascade}, in the first iteration, the model can only select partial object regions. After updating the learnable embeddings with the selected features, it expands to cover more potential object regions. Through multiple iterations, the model is able to propose most potential object regions.

\noindent{\bf Sparse Image-Aware Adapter.} In the Sparse Image-Aware Adapter (SIA), we dynamically adapt the learnable embedding using visual features, enabling it to propose objects from unseen categories. To evaluate the effectiveness of this module, we visualize the regions selected by the learnable embedding before and after the SIA update. As shown in \cref{fig:moe}, before the update, the learnable embedding assigns high attention to background regions. If these regions are fed into the CSP module, the model tends to propose more background regions. In contrast, after being updated with visual features, the learnable embedding focuses on object regions and suppresses background distractions, underscoring the necessity of the SIA.

\noindent{\bf Centerness-Guided Query Selection.} 
The core idea of CG-QS is that image queries located near an object's center tend to generate more accurate proposals than those near the object boundary. As shown in \cref{fig:center}, when the model selects a center-area query (the red star), it produces a precise bounding box. In contrast, when the module selects queries from boundary regions, it typically results in notable localization errors. Motivated by this observation, we introduce a center loss to encourage the model to prioritize queries closer to object centers. To evaluate the effectiveness of the CG-QS strategy, we visualize the selected image queries, as shown in \cref{fig:cencervisual}. After adding the CG-QS module, our model prefers to select queries at center locations, confirming the efficacy of the CG-QS strategy.

\section{Conclusion}
In this paper, we propose the Prompt-Free Universal Region Proposal Network~(\ourmodel), which aims to address a critical limitation in computer vision: the task of proposing arbitrary potential objects typically depends on external prompts~(\eg, text descriptions or visual cues). To mitigate this limitation, \ourmodel~introduces a learnable embedding serving as a proxy for text embeddings, enabling prompt-free arbitrary object proposal.  We propose the Sparse Image-Aware Adapter and Cascade Self-Prompt modules, which enhance the model's localization capability as the similarity among visual embeddings is typically greater than that between the learnable embedding and visual embeddings.
We further present the Centerness-Guided Query Selection module, which incorporates centerness and classification scores to select more appropriate queries for subsequent stages. 
Extensive experiments demonstrate our \ourmodel's superiority in zero-shot cross-domain object proposal, providing valuable insights for future research.

\clearpage
\setcounter{page}{1}
\maketitlesupplementary
\setcounter{section}{0}

\section{Ablation Study on \textit{k}}

In \cref{sec:sia}, we introduce sparsity to adaptively select the top-$k$ informative feature maps for updating the learnable embedding. In this section, we ablate the choice of $k$ in the Sparse Image-Aware Adapter module on the CD-FSOD benchmark to examine the effect of sparsity.

As shown in \cref{tab:k-ablation}, \ourmodel~achieves the best overall performance when $k = 2$, thereby choosing $k = 2$ as the default setting in our framework. Increasing $k$ introduces more redundant feature maps and slightly degrades performance, while too small a $k$ limits the available contextual information. This demonstrates that moderate sparsity offers the best trade-off.

\begin{table}[htbp]
\small
\centering
\caption{Ablation study of parameter $k$ in the Sparse Image-Aware Adapter module on the CD-FSOD benchmark. Best results are highlighted in \textbf{bold}.}
\label{tab:k-ablation}
\begin{tabular}{lcccccc}
\toprule
\rowcolor{HeaderGray}
$k$ & AR$_{100}$ & AR$_{300}$ & AR$_{900}$ & AR$_s$ & AR$_m$ & AR$_l$ \\
\midrule
1 & 60.3 & 64.7 & 67.4 & \textbf{44.2} & 60.0 & 78.8 \\
\rowcolor{LightBlueRow}
2 & \textbf{60.7} & \textbf{65.3} & \textbf{68.2} & 38.5 & \textbf{61.9} & \textbf{80.3} \\
3 & 59.9 & 64.4 & 67.3 & 41.5 & 59.0 & 79.6 \\
\rowcolor{LightBlueRow}
4 & 59.7 & 64.4 & 67.3 & 41.5 & 59.0 & 79.6 \\
\bottomrule
\end{tabular}
\end{table}

\section{Ablation Study on Objective Loss}
In our objective loss function, we introduce a hyperparameter $\lambda$ to control the contribution of the centerness loss to the overall loss. To determine an appropriate setting, we conduct an ablation study on 
$\lambda$. As shown in Table~\ref{tab:lam_ablation}, when 
$\lambda$ is too small, the model does not learn to select queries located in the center regions. In contrast, when $\lambda$ is too large, the centerness loss dominates optimization and negatively affects regression performance. The best performance is achieved when $\lambda$ = 5.

\begin{table}[htbp]
\small
\centering
\caption{Ablation study of parameter $\lambda$ in the Objective Loss on CD-FSOD benchmark. The best results are highlighted in \textbf{bold}.}
\label{tab:lam_ablation}
\begin{tabular}{lcccccc}
\toprule
\rowcolor{HeaderGray}
$k$ & AR$_{100}$ & AR$_{300}$ & AR$_{900}$ & AR$_s$ & AR$_m$ & AR$_l$ \\
\midrule
1 & 60.5&	65.1&	68.1&	39.6&	59.8&	79.4\\
\rowcolor{LightBlueRow}
3 & 60.6&	65.1&	68.1&	39.6&	60.3&	79.5\\
5 & \textbf{60.7} & \textbf{65.3} & \textbf{68.2} & 38.5 & \textbf{61.9} & \textbf{80.3} \\
\rowcolor{LightBlueRow}
7&58.5&	63.1&	66.3&	37.1&	58.3&	79.2\\
9 & 59.6&	64.3&	67.4&	39.7&	59.6&	78.9\\
\bottomrule
\end{tabular}
\end{table}
\section{Efficacy of Self-Prompt}

\begin{figure}[htbp]
    \centering
    \begin{subfigure}{0.45\linewidth}
      \includegraphics[width=\linewidth]{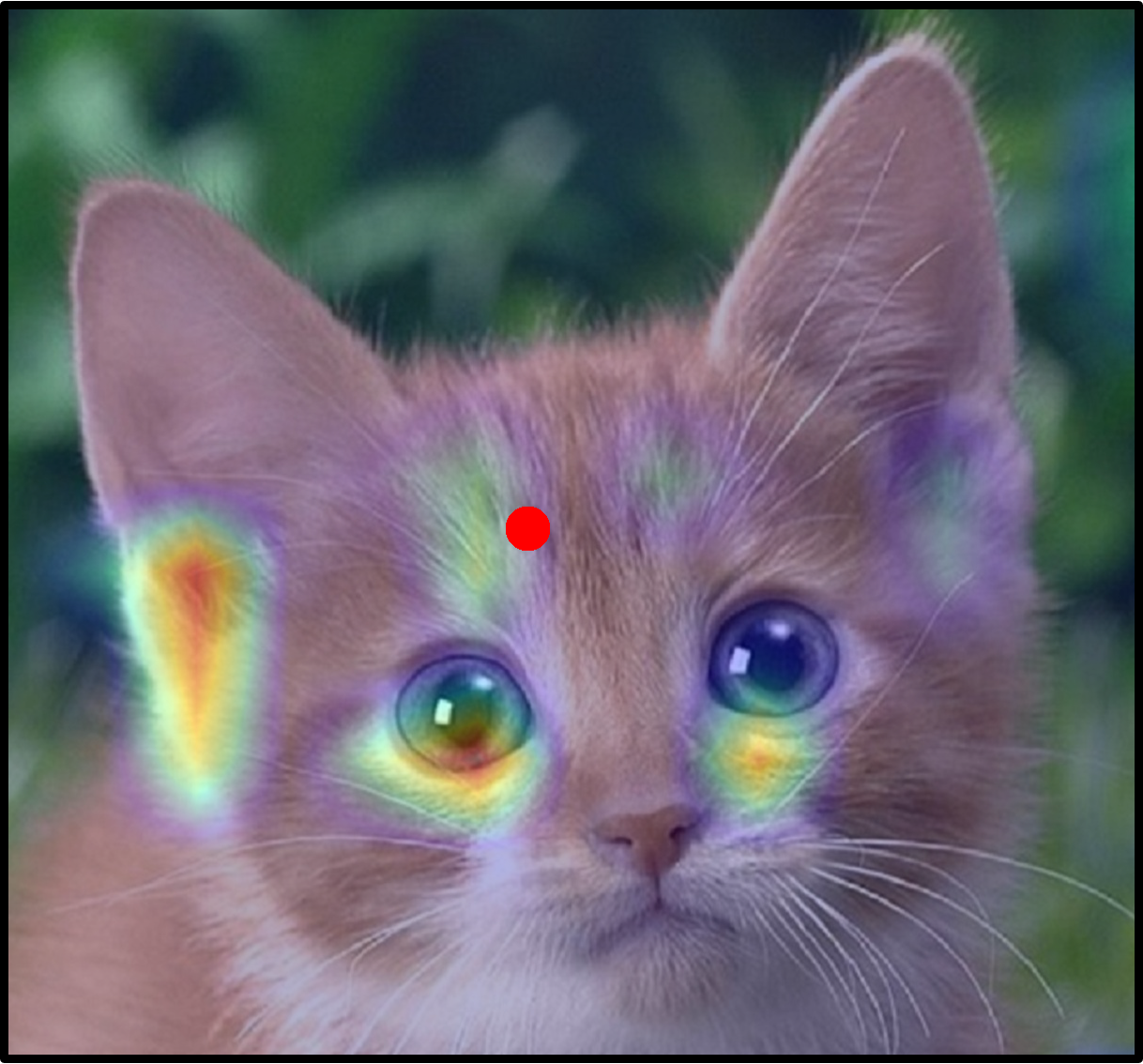}
      \caption{Cosine similarity heatmap between the 4th-level feature at the red point and the 4th-level visual embedding.}
      \label{fig:csp-compare-a}
    \end{subfigure}
    \hfill
    \begin{subfigure}{0.45\linewidth}
      \includegraphics[width=\linewidth]{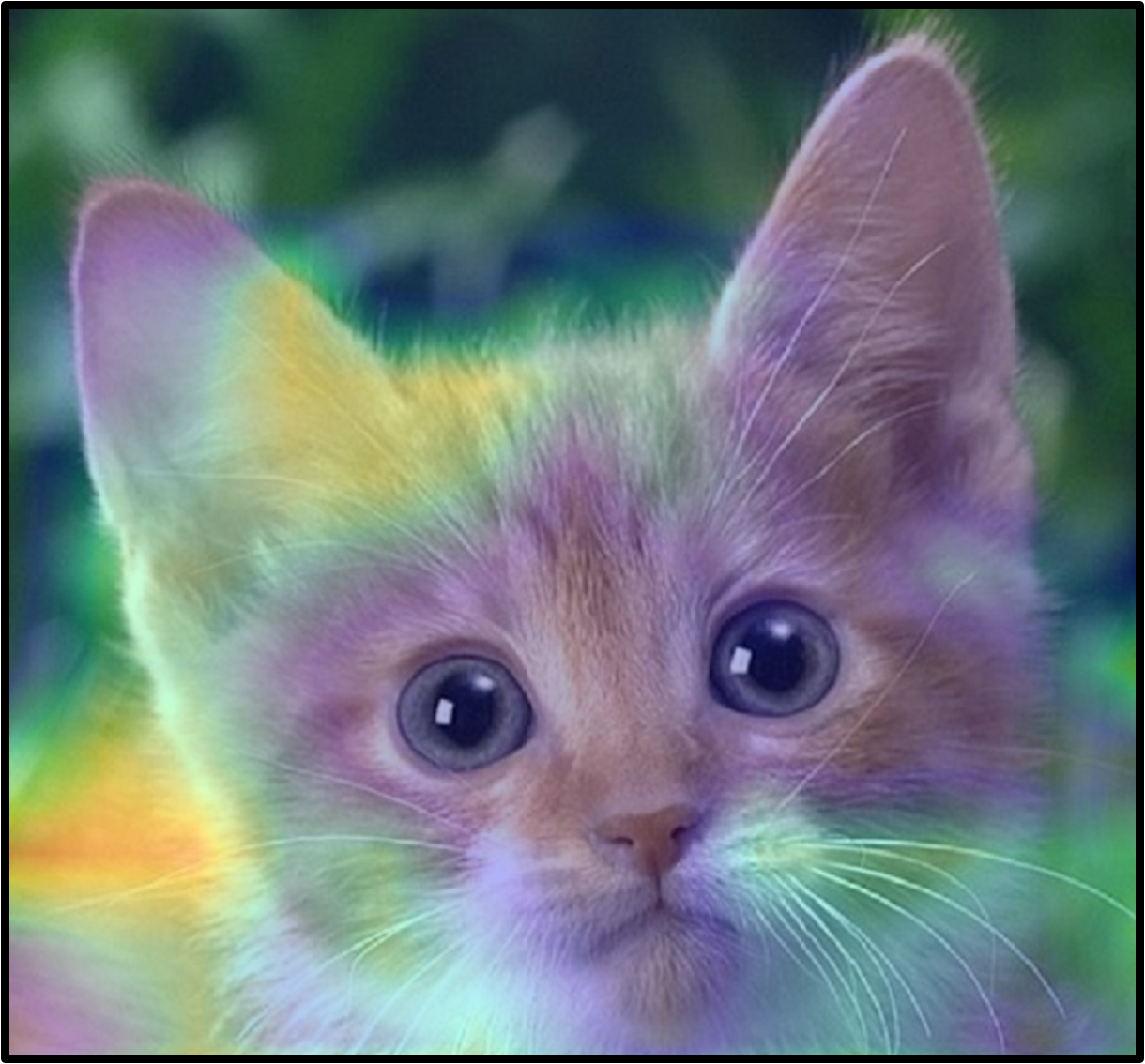}
      \caption{Cosine similarity heatmap between the learnable embedding $\Tilde{F}^T$ and the 4th-level visual embedding.}
      \label{fig:csp-compare-b}
    \end{subfigure}
    \caption{Comparison of feature localization between object-internal features and the learnable embedding $\Tilde{F}^T$. The similarity map in \cref{fig:csp-compare-a} shows that the 4th-level feature at the red point focuses on the object region with high semantic consistency, while the learnable embedding $\Tilde{F}^T$ in \cref{fig:csp-compare-b} produces a more diffused response, indicating weaker object correspondence.}
    \label{fig:csp-compare}
\end{figure}

As shown in \cref{fig:csp-compare}, the object-internal feature in \cref{fig:csp-compare-a} focuses on the object region with high semantic consistency, whereas the learnable embedding in \cref{fig:csp-compare-b} yields a more diffused response.

These observations indicate that object-internal features exhibit stronger localization capability than the learnable embedding. Therefore, in \cref{sec:csp}, we leverage multi-level feature maps to update the learnable embedding, further enhancing its ability to localize objects based on internal visual cues.

\section{Latency and Efficiency Analysis}

The proposed iterative \textit{Cascade Self-Prompt (CSP)} strategy introduces negligible latency overhead. As shown in \cref{tab:latency_iter}, an increase in the number of CSP iterations from 1 to 3 yields consistent performance gains, accompanied by only a marginal rise in inference time ($\sim$4.6 ms).

\begin{table}[htbp]
\centering
\caption{Latency and performance analysis of different CSP iterations on the CD-FSOD benchmark.}
\label{tab:latency_iter}
\resizebox{\linewidth}{!}{
    \small
    \renewcommand{\arraystretch}{1.1}
    \begin{tabular}{lccccccc}
    \toprule
    \rowcolor{HeaderGray}
    Iteration & $AR_{100}$ & $AR_{300}$ & $AR_{900}$ & $AR_{s}$ & $AR_{m}$ & $AR_{l}$ & ms/img \\
    \midrule
    Iter 1 & 59.6 & 63.1 & 65.4 & 35.7 & 57.8 & 77.7 & \textbf{214.3} \\
    \rowcolor{LightBlueRow}
    Iter 2 & 59.9 & 63.7 & 65.9 & 36.7 & 57.9 & 78.2 & 216.7\\
    Iter 3 & \textbf{60.7} & \textbf{65.3} & \textbf{68.2} & \textbf{38.5} & \textbf{61.9} & \textbf{80.3} & 218.9 \\
    \bottomrule
    \end{tabular}
}
\end{table}

Furthermore, the proposed method exhibits high flexibility and can be seamlessly integrated with lightweight detectors to function as a real-time, high-performance region proposal network (RPN~\cite{ren2016faster}). As demonstrated in \cref{tab:latency_context}, the integration of the proposed approach with YOLO-World~\cite{Cheng2024YOLOWorld} achieves competitive performance while preserving inference speeds comparable to those of conventional RPNs.

\begin{table}[htbp]
\centering
\caption{Efficiency comparison of PF-RPN integrated with different detectors.}
\label{tab:latency_context}
\resizebox{\linewidth}{!}{
    \small
    \begin{tabular}{lccc|cc}
    \toprule
    \rowcolor{HeaderGray}
    \textbf{Metric} & GLIP~\cite{liu2024grounding} & CasRPN~\cite{vu2019cascade} & RPN~\cite{ren2016faster} & \textbf{PF-RPN(GDINO)} & \textbf{PF-RPN(YWorld)}\\
    \midrule
    \textbf{$AR_{100}$}  & 47.6 & 45.8 & 32.0 & \textbf{60.7} & 52.3 \\
    \rowcolor{LightBlueRow}
    \textbf{FPS} & 5.5 & 24.8 & 27.8 & 4.6 & \textbf{25.1} \\
    \bottomrule
    \end{tabular}
}
\end{table}

\section{Analysis of False Positives}

The RPN is designed to detect all potential objects, a process that inevitably leads to the proposal of task-irrelevant regions, thereby generating false positives (FPs). In comparison to existing RPNs, the proposed PF-RPN assigns higher confidence scores to true positive object candidates while effectively suppressing irrelevant regions. As indicated in \cref{tab:baseline_comparison}, the proposed method achieves more substantial improvements in AP and a lower number of false positives when restricted to 100 proposals compared to the 300-proposal setting. This result demonstrates the capacity of the proposed approach to prioritize high-quality candidates and mitigate redundant false positives.

\begin{table}[htbp]
    \centering
    \caption{Analysis of false positives on different baselines under varying top proposal settings.}
    \label{tab:baseline_comparison}
    \resizebox{\linewidth}{!}{
        \renewcommand{\arraystretch}{1.2}
        \begin{tabular}{l|cc|cc|cc|cc}
            \toprule
            \rowcolor{HeaderGray}
             & \multicolumn{2}{c|}{DeViT~\cite{xu2023devit}} & \multicolumn{2}{c|}{\textbf{DeViT + Ours}} & \multicolumn{2}{c|}{CD-ViTO~\cite{fu2024cross}} & \multicolumn{2}{c}{\textbf{CD-ViTO + Ours}} \\
            \rowcolor{HeaderGray} \multirow{-2}{*}{Metric}
             & Top 100 & Top 300 & Top 100 & Top 300 & Top 100 & Top 300 & Top 100 & Top 300 \\
            \midrule
            FP ($\downarrow$) & 18.2 & 21.7 & 17.1 & 18.3 & 15.3 & 20.2 & \textbf{14.6} & 17.9 \\
            \rowcolor{LightBlueRow}
            AP ($\uparrow$) & 32.3 & 33.4 & \textbf{37.3 (+5.0)} & 37.1 (+3.7) & 26.8 & 29.0 & 34.2 (+7.4) & 35.1 (+6.1) \\
            \bottomrule
        \end{tabular}
    }
\end{table}

\section{Dependence on Base Detectors}

The proposed method exhibits strong extensibility and can be effectively integrated with various base detectors. As presented in \cref{tab:mmgddino}, the proposed approach derives direct benefits from more powerful base detectors, demonstrating steady performance improvements as the capacity of the base model increases.

\begin{table}[htbp]
\centering
\caption{Performance comparison of integrating PF-RPN with stronger base models (MMGrounding DINO~\cite{zhao2024open}).}
\label{tab:mmgddino}
\resizebox{\linewidth}{!}{
\small
\begin{tabular}{lcccccc} 
\toprule
\rowcolor{HeaderGray}
Method & {$AR_{100}$} & {$AR_{300}$} & {$AR_{900}$} & {$AR_{s}$} & {$AR_{m}$} & {$AR_{l}$}\\
\midrule
MMGDINO-B & 72.4 & 73.5 & 73.9 & \textbf{46.4} & 66.7 & 78.7\\
\rowcolor{OursHighlight}
+ Ours  & \textbf{76.3} & \textbf{78.9} & \textbf{79.8} & 45.7 & \textbf{75.2} & \textbf{86.3}\\
\midrule
MMGDINO-L & 73.6 & 74.2 & 74.5 & 48.3 & 69.8 & 79.4 \\
\rowcolor{OursHighlight}
+ Ours & \textbf{77.4} & \textbf{79.1} & \textbf{80.1} & \textbf{48.5} & \textbf{78.4} & \textbf{86.5}\\
\bottomrule
\end{tabular}
}
\end{table}

\section{Comparison with Previous Prompt-Free Methods}

We compare PF-RPN with representative open-source prompt-free methods, GenerateU~\cite{lin2024generative} and Open-Det~\cite{caoopen}. We do not include CapDet~\cite{long2023capdet} and DetCLIPv3~\cite{yao2024detclipv3} due to the unavailability of their official code. As presented in \cref{tab:prompt_free_comparison}, PF-RPN surpasses GenerateU by \textbf{+13.0 $AR_{100}$} on CD-FSOD, while reducing VRAM usage by \textbf{95\%} and accelerating inference by nearly \textbf{20$\times$}. Note that PF-RPN is also faster than the baseline GDINO~\cite{liu2024grounding}, primarily due to the removal of the computationally expensive text encoder.

\begin{table}[htbp]
    \centering
    \caption{Comparison with open-source prompt-free methods regarding performance and efficiency.}
    \label{tab:prompt_free_comparison}
    \resizebox{\linewidth}{!}{
        \renewcommand{\arraystretch}{1.1}
        \begin{tabular}{cl|ccc|ccc|cc}
            \toprule
            \rowcolor{HeaderGray}
             &  & \multicolumn{3}{c|}{Average Recall (AR)} & \multicolumn{3}{c|}{Scale ($AR$)} & \multicolumn{2}{c}{Efficiency} \\
            \rowcolor{HeaderGray}
             \multirow{-2}{*}{Benchmark} & \multirow{-2}{*}{Method} & $AR_{100}$ & $AR_{300}$ & $AR_{900}$ & $S$ & $M$ & $L$ & FPS & VRAM \\
            \midrule
            
             & GDINO~\cite{liu2024grounding} & 54.7 & 57.8 & 61.6 & 34.1 & 49.3 & 67.0 & 3.3 & 0.9G \\
            \rowcolor{LightBlueRow} \cellcolor{white} & GenerateU~\cite{lin2024generative} & 47.7 & 54.1 & 55.7 & 28.1 & 48.3 & 69.4 & 0.22 & 12.2G \\
             & Open-Det~\cite{caoopen} & 36.6 & 46.3 & 54.3 & 28.2 & 45.3 & 67.7 & 0.15 & 30.7G \\
            \cmidrule{2-10}
            \rowcolor{OursHighlight} \cellcolor{white} \multirow{-4}{*}{\rotatebox[origin=c]{90}{\textbf{CD-FSOD}}} & \textbf{PF-RPN (Ours)} & \textbf{60.7} & \textbf{65.3} & \textbf{68.2} & \textbf{38.5} & \textbf{61.9} & \textbf{80.3} & \textbf{4.6} & \textbf{0.5G} \\
            \midrule
            
             & GDINO~\cite{liu2024grounding}  & 69.1 & 70.9 & 72.4 & 40.8 & 64.6 & 78.4 & 3.3 & 0.9G \\
            \rowcolor{LightBlueRow} \cellcolor{white} & GenerateU~\cite{lin2024generative}  & 67.3 & 71.5 & 72.2 & 32.8 & 63.1 & 80.0 & 0.22 & 12.2G \\
             & Open-Det~\cite{caoopen}  & 53.9 & 62.9 & 69.1 & 27.7 & 59.8 & 76.6 & 0.15 & 30.7G \\
            \cmidrule{2-10}
            \rowcolor{OursHighlight} \cellcolor{white} \multirow{-4}{*}{\rotatebox[origin=c]{90}{\textbf{ODinW13}}} & \textbf{PF-RPN (Ours)} & \textbf{76.5} & \textbf{78.6} & \textbf{79.8} & \textbf{45.4} & \textbf{71.9} & \textbf{85.8} & \textbf{4.6} & \textbf{0.5G} \\
            \bottomrule
        \end{tabular}
    }
\end{table}

\section{Detailed Experimental Results on All 19 Datasets}

\begin{figure}[htbp]
    \centering
    \includegraphics[width=\linewidth]{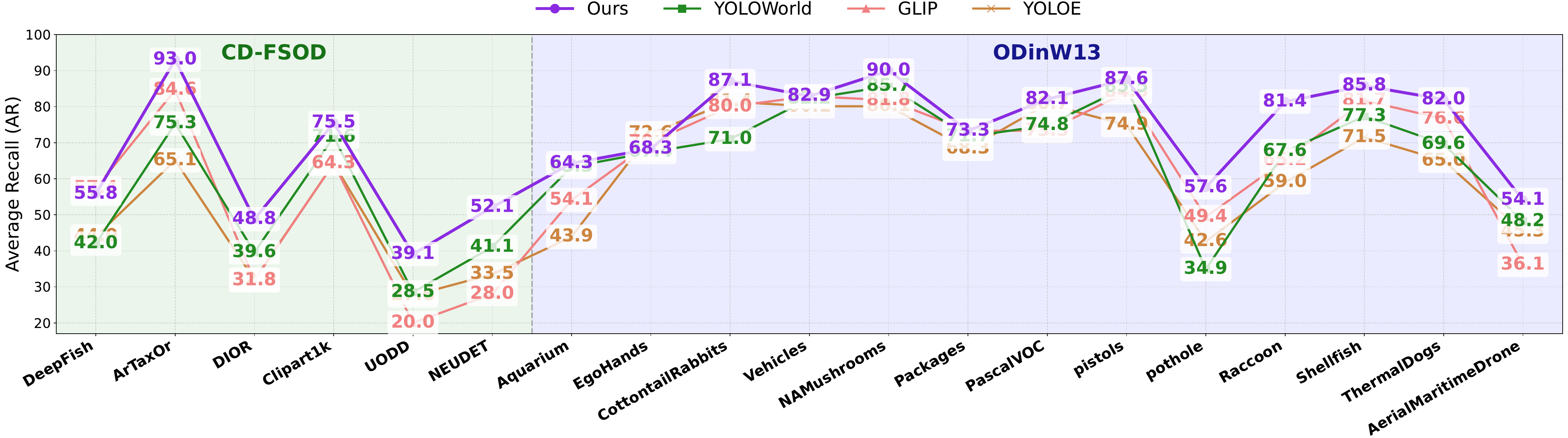}
    \caption{Detailed performance ($AR$) trends on all 19 target datasets compared to alternative methods.}          
    \label{fig:detailed_results}
\end{figure}

Comprehensive performance metrics of the proposed method are reported on various datasets across both the CD-FSOD benchmark and the ODinW13 benchmark. Additionally, \cref{fig:detailed_results} presents a visual line-chart comparison of these results against those of existing baseline methods.

\clearpage  

\section*{Acknowledgements}

This work is supported in part by the National Natural Science Foundation of China (62192783, 62276128, 62406140), Young Elite Scientists Sponsorship Program by China Association for Science and Technology (2023QNRC001), the Key Research and Development Program of Jiangsu Province under Grant (BE2023019) and Jiangsu Natural Science Foundation under Grant (BK20221441, BK20241200). The authors would like to thank Huawei Ascend Cloud Ecological Development Project for the support of Ascend 910 processors.

{
    \small
    \bibliographystyle{ieeenat_fullname}
    \bibliography{main}
}


\end{document}